\documentclass{article}

\usepackage{microtype}
\usepackage{graphicx}
\usepackage{subfigure}
\usepackage{booktabs} % for professional tables
\usepackage{amsmath}

\usepackage{hyperref}

\usepackage[accepted]{icml2020}

\icmltitlerunning{FiniteNet: A Fully Convolutional LSTM Network Architecture for Time-Dependent Partial Differential Equations}

\begin{document}

\twocolumn[
\icmltitle{FiniteNet: A Fully Convolutional LSTM Network\\
           Architecture for Time-Dependent Partial Differential Equations}



\icmlsetsymbol{equal}{*}

\begin{icmlauthorlist}
\icmlauthor{Ben Stevens}{caltech}
\icmlauthor{Tim Colonius}{caltech}
\end{icmlauthorlist}

\icmlaffiliation{caltech}{Department of Mechanical and Civil Engineering, California Institute of Technology, Pasadena, CA, United States}

\icmlcorrespondingauthor{Ben Stevens}{bstevens@caltech.edu}

\icmlkeywords{Machine Learning, ICML}

\vskip 0.3in
]



\printAffiliationsAndNotice{\icmlEqualContribution} % otherwise use the standard text.

\begin{abstract}
In this work, we present a machine learning approach for reducing the error when numerically solving time-dependent partial differential equations (PDE). We use a fully convolutional LSTM network to exploit the spatiotemporal dynamics of PDEs. The neural network serves to enhance finite-difference and finite-volume methods (FDM/FVM) that are commonly used to solve PDEs, allowing us to maintain guarantees on the order of convergence of our method. We train the network on simulation data, and show that our network can reduce error by a factor of 2 to 3 compared to the baseline algorithms. We demonstrate our method on three PDEs that each feature qualitatively different dynamics. We look at the linear advection equation, which propagates its initial conditions at a constant speed, the inviscid Burgers' equation, which develops shockwaves, and the Kuramoto-Sivashinsky (KS) equation, which is chaotic.
\end{abstract}
%See https://www.overleaf.com/8277387563hrvqskpdckbv for an appendix I have moved some things to

\section{Introduction}
\subsection{Motivation}
Partial differential equations (PDE) are fundamental to many areas of physics, such as fluid mechanics, electromagnetism, and quantum mechanics \cite{sommerfeld1949partial}. All classical physics models are approximations obtained by coarse-graining the true quantum mechanical behavior of matter. For example, the Navier-Stokes equations are obtained by treating a fluid as a continuum; despite this, the equations are still too computationally expensive to solve for most practical problems \cite{ishihara2009study}, and so they are further coarse-grained by methods such as Large-Eddy Simulation \cite{halpern1993large} or Reynolds Averaged Navier Stokes \cite{chen1990solutions} to infer sub-grid behavior. Each of these coarse-grained models save multiple orders of magnitude of computational expense \cite{drikakis2006turbulent, wilcox1998turbulence}. However, these approaches do not lead to models that are generally applicable \cite{spalart2010reflections}.%, and sometimes produce results that are untrustworthy when solving hard problems, such as transition from laminar to turbulent flow \cite{zhiyin2015large}. 

\begin{figure}[h!]\label{DiffAx}
\centering
\includegraphics[width=0.4\textwidth]{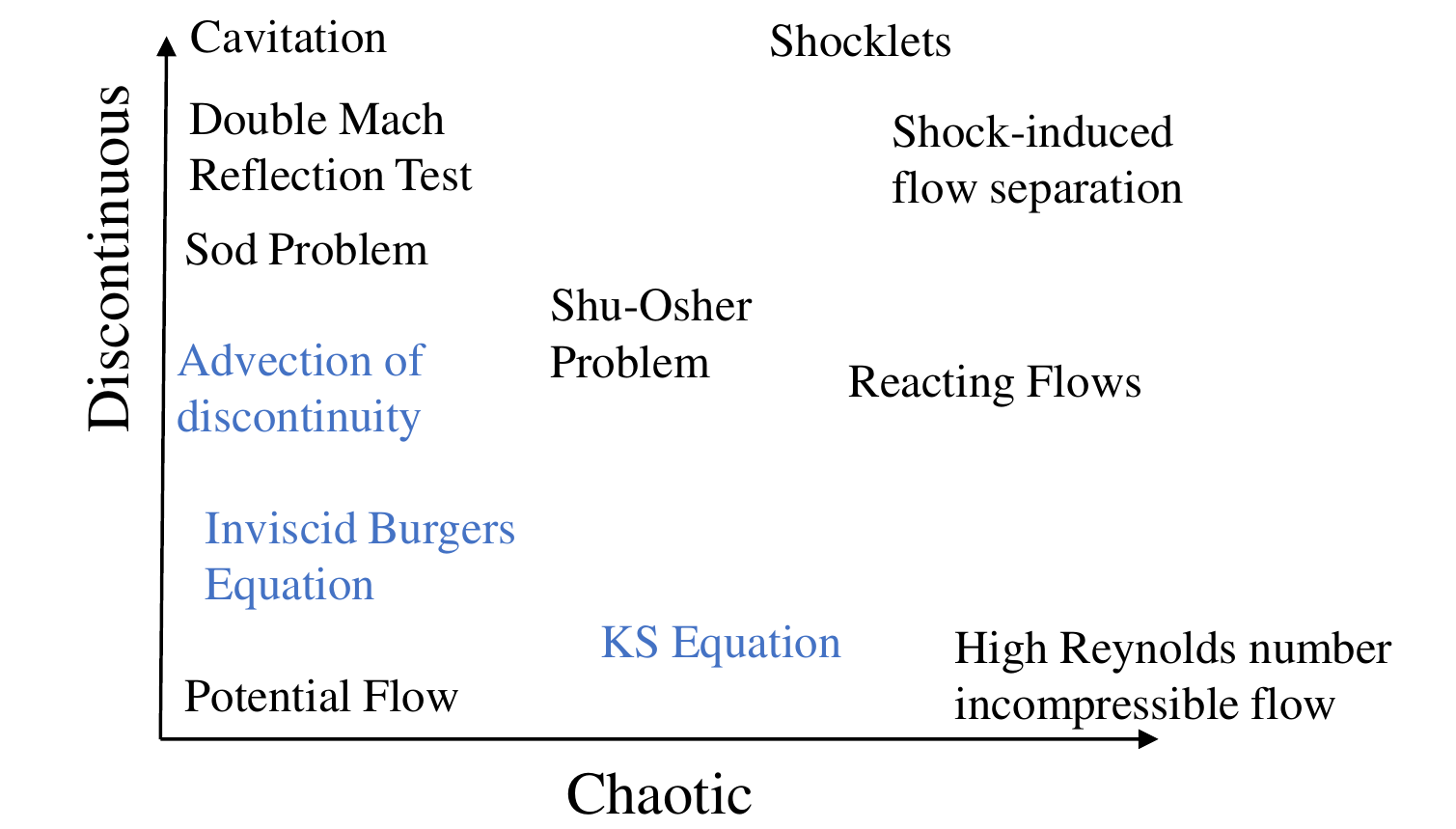}
\caption[batch1c]{Problems with discontinuous and chaotic behavior. Problems in blue are examined in our experiments.}
\end{figure}

PDEs give rise to vastly different solution structures in different problems \cite{sommerfeld1949partial}. Hence, a solution approach that works well for one equation will not be generally applicable, as different difficulties can cause different methods to fail. The two major difficulties we examine in this paper are PDEs with chaotic dynamics, and PDEs with discontinuous solutions. PDEs with chaotic dynamics are challenging because small errors will grow quickly in time, causing the numerical solution to diverge from the true solution if the solver is not accurate enough \cite{strogatz2001nonlinear}. Typically, high-order numerical methods are used to solve these problems as they offer the best asymptotic error bounds \cite{deville2002high}. PDEs with discontinuous solutions are difficult to solve because high-order methods lead to Gibbs phenomena near discontinuities \cite{gottlieb1997gibbs}, which can lead to numerical instabilities. High-order methods are derived with the assumption that the solution is smooth\cite{leveque2007finite}, but no method can achieve better than first-order accuracy in the presence of a discontinuity \cite{leveque2002finite}. % so order of accuracy is not prioritized \cite{harten1983high}. 
Hence, we can see methods used to solve turbulence and discontinuities are at odds with each other, which makes it especially challenging to simulate problems with both of these issues. In \ref{DiffAx}, we present a qualitative overview of problems in fluid mechanics that involve turbulent and/or discontinuous solutions.

%Other difficulties can arise in these problems such as multiphysics behavior (magnetohydrodynamics \cite{davidson2002introduction, devore1991flux, gammie2003harm}, combustion \cite{peters2001turbulent, linan1993fundamental, veynante2002turbulent}, fluid-structure interaction \cite{bungartz2006fluid, bazilevs2008isogeometric, le2001fluid,dowell2001modeling}), complex geometries \cite{kim2001immersed,mahesh2004numerical,mohd1997simulations}, and many other challenges \cite{borggaard2004computational, gustafsson2010implementation, funaki1995scaling, ricardez2011current, givoli1991non}.

Numerical methods such as FDM and FVM are often developed without knowledge of the specific equation that they will be used to solve, and instead opt to maximize the rate of convergence \cite{leveque2007finite}. However, the dynamics of PDEs greatly influence the structure of their solutions. This information should be used to maximize the performance of the numerical methods used to simulate these equations. Machine learning is the natural tool to hybridize this information with traditional numerical methods, as detailed experimental/simulation data that contains this behavior is available and can be used to improve these numerical methods. Furthermore, machine learning techniques for capturing temporal behavior such as LSTMs \cite{hochreiter1997long, hochreiter1997lstm, gers1999learning}, and spatially local behavior such as CNNs \cite{fukushima1980neocognitron, lecun1989backpropagation, lecun1998gradient}, have undergone significant development. This makes machine learning highly effective for modeling problems with these structures, both of which occur in time-dependent PDEs.

\subsection{Related Work}
Research intersecting PDEs with ML can be divided into two main goals: discovering PDEs from data, and using ML to better solve PDEs \cite{raissi2019physics}. Although our paper focuses on solving PDEs, we will briefly discuss efforts to discover them from data. Brunton et al. \yrcite{brunton2016discovering} learned dynamical systems by applying lasso regression to a library of functions. Many papers used LSTMs or other sequence models to learn dynamical systems \cite{hagge2017solving, yu2017long, vlachas2018data, wan2018data, li2017diffusion}. Other papers learned coarse-graining models by following a system identification approach \cite{ling2016reynolds}.%An important quality of PDEs is the presence of spatiotemporal dynamics. Many ML approaches have been developed to make inferences about systems with this characteristic. Shi et al. \yrcite{xingjian2015convolutional} used a convolutional LSTM to predict short term rainfall based on radar maps. Mohan et al. \yrcite{mohan2019compressed} developed a deep learning framework called compressed convolutional LSTM to reduce the dimensionality of turbulence.

Solving PDEs with machine learning can be further broken up into two main areas: using data to develop better solvers, and parameterizing solutions to PDEs as a neural network and learning weights to minimize the pointwise error of the PDE \cite{lagaris1998artificial}. This second idea has been further developed \cite{rudd2013solving} to solve problems such as high-dimensional PDEs \cite{sirignano2018dgm}. %PDE solvers can entail several different things. 
Hsieh et al. \yrcite{hsieh2019learning} learned iterative PDE solvers for time-independent problems that are solved via matrix inversion. In our work, we focus on time-dependent problems that are solved via time-stepping, though both involve discretization. There have been numerous efforts to develop discretization methods tailored to specific types of dynamics. In fluid mechanics, much work has gone into developing shock capturing methods, spatial discretization algorithms specially suited towards solving PDEs with discontinuous solutions \cite{harten1983high}. This strategy can be viewed as a form of coarse-graining, as the true equations have viscous terms that prevent discontinuities from forming. They instead form very steep but continuous features, but these require a very fine grid to fully resolve. There have also been machine learning strategies that develop equation-specific spatial discretization schemes. The idea of embedding machine learning models into spatial discretization methods for the purpose of coarse-graining was introduced by Bar-Sinai et al. \yrcite{bar2019learning}. In their paper, they trained a neural network to interpolate over the solution of a PDE to more accurately predict the solution at the next timestep. PDE-Net followed a similar approach \cite{long2017pde, long2019pde} in learning derivatives while also learning the PDE from data.

\subsection{Our Contribution}
We try to simultaneously exploit the spatial and temporal structure of PDEs through the convolutional LSTM architecture of our neural network, while maintaining the computational structures used in FVM/FDM. While some papers use convolutional LSTMs or related architectures to predict future values of time-series that also possess the property of spatial locality \cite{mohan2019compressed,xingjian2015convolutional}, none exploit the FVM/FDM structure that naturally discretizate PDEs. Other papers have developed coarse-graining models by learning from data and have embedded these into CNNs \cite{bar2019learning}, but none simultaneously used an LSTM structure to also exploit the temporal structure of PDEs. We also present a novel training approach of directly minimizing simulation error over long time-horizons by building upon ideas developed for PDE-Net \cite{long2017pde}. Our approach aims to push the field of data-driven scientific computing forward by combining these ideas in a sufficiently generic and efficient framework to permit extensions to many problems.%to fully utilize the structure of PDEs to make our numerical methods as efficient as possible.

\section{Numerical Methods Background}
\subsection{Strong Stability Preserving Runge-Kutta Methods}
Runge-Kutta methods are a family of numerical integration methods used to advance a differential equation $\frac{\partial u}{\partial t}=L(u)$ forward in time. In this paper, we use SSPRK3 \cite{gottlieb1998total}

\begin{equation}
    \begin{split}
u^{(1)} = u^{(n)} + \Delta tL(u^{(n)}),\\
u^{(2)} = \frac{3}{4}u^{(n)} + \frac{1}{4}u^{(1)} + \frac{1}{4}\Delta tL(u^{(1)}),\\
u^{(n+1)} = \frac{1}{3}u^{(n)} + \frac{2}{3}u^{(2)} + \frac{2}{3}\Delta tL(u^{(2)}),
    \end{split}
\end{equation}

a three-step, third-order accurate method that preserves the total variation diminishing (TVD) property $TV(u(t_{i+1})) \leq TV(u(t_{i}))$ of explicit Euler, where total variation is defined as
\begin{equation}
    TV(u) = \sum_{i=1}^N|u(x_i)-u(x_{i-1})|.
\end{equation}
A TVD method prevents the time-stepping method from adding spurious oscillations to the solution, which is a major concern for PDEs with discontinuous solutions, as they can lead to instabilities that crash the simulation in nonlinear PDEs such as the inviscid Burgers' equation.

\subsection{Finite Difference Method}
One spatial discretization method we combine with machine learning is FDM. In this method, derivatives of the solution are approximated as a weighted combination of local values of the solution. For example, one could approximate the first derivative of a function as

\begin{equation}
    \label{FDM_eq}
    \frac{\partial u}{\partial x} \approx \frac{1}{\Delta x}\sum_{i=-1}^{1}c_iu_i=\frac{u_1-u_{-1}}{2\Delta x}
\end{equation}

by using coefficients $c_{-1}=-\frac{1}{2}$, $c_{0}=0$, and $c_{1}=\frac{1}{2}$. We use the facts that the bounds on $i$ can be expanded and the values of $c_i$ are not fully constrained to allow a neural network to chose these coefficient values based on the local solution as $c_{a:b}=f(u_{a:b})$.

\subsection{Finite Volume Method}
We also consider equations that are more natually solved using FVM, a spatial discretization method similar to FDM. FVM offers the advantage of improved stability \cite{bar2019learning}, and is easier to extend to unstructured meshes \cite{chen2003unstructured}.
\subsubsection{Example}
Consider the scalar conservation law
\begin{equation}
    \frac{\partial u}{\partial t} +\frac{\partial f(u)}{\partial x} =0
\end{equation}
One can split the $x$-domain into cells of width $\Delta x$ and average over them as
\begin{equation}
    \frac{1}{\Delta x}\int_{x_i}^{x_i+\Delta x}\frac{\partial u}{\partial t} +\frac{\partial f(u)}{\partial x}dx =0
\end{equation}
to get 
\begin{equation}
    \frac{\partial \Bar{u}_i}{\partial t} +\frac{f(u(x_i+\Delta x))-f(u(x_i))}{\Delta x} =0.
\end{equation}
Note that this equation is still exact. The cell average values $\Bar{u}$ are tracked and interpolated to find $u$ locally as 
\begin{equation}
    u(x_i) \approx \sum_{j=a}^{b}c_j\Bar{u}_j.
\end{equation}
Once again, the coefficients $c_j$ are not fully constrained and can be determined using machine learning.

\section{PDE Background}
%\tim{You could eliminate the subsections here to save space if needed} \ben{the paper is within the page limit so I will leave it for now}

\begin{figure*}[h!]\label{All3Solutions}
\centering
\includegraphics[width=1\textwidth]{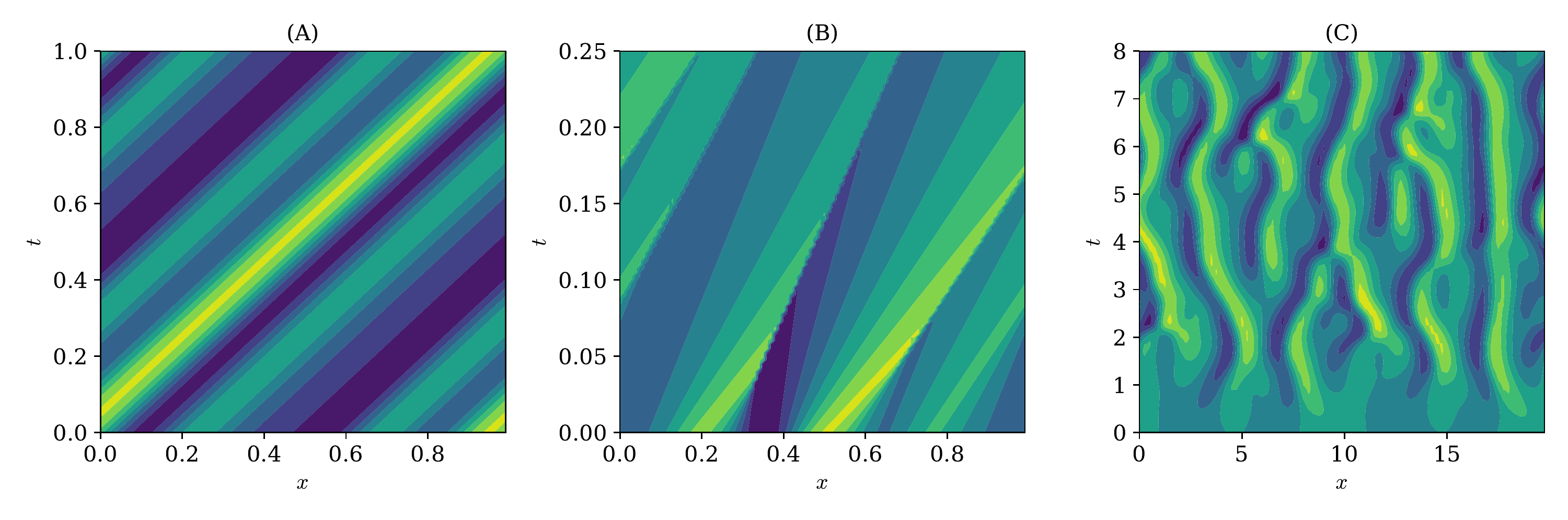}
\caption[batch1c]{Example solution of (A) linear advection, (B) inviscid Burgers', and (C) KS equations}
\end{figure*}
\subsection{Advection Equation}
The linear advection equation is written as
\begin{equation}
    \frac{\partial u}{\partial t} + a\frac{\partial u}{\partial x} = 0.
\end{equation}
It possesses solutions that are translations of the initial conditions at wavespeed $a$, i.e. for an initial condition of $u(x,0)=f(x)$, the exact solution is $u(x,t)=f(x-at)$. In terms of coarse-graining, the sub-grid behavior has no influence on the solution at subsequent times. Once discretization occurs, no inference can be made about what happens between grid points, as this is fully determined by the initial condition rather than the dynamics of the equation. When this equation is solved with a discontinuous initial condition, it serves as a toy problem for advecting different materials in a multicomponent/multiphase flow. Numerical error causes the discontinuity to become smeared out during the simulation.

\subsection{Inviscid Burgers' Equation}
The inviscid Burgers' equation is written in non-conservative form as
\begin{equation}
    \frac{\partial u}{\partial t} + u\frac{\partial u}{\partial x} = 0,
\end{equation}
and is used to model wave-breaking. The nonlinear term causes shockwaves to form in finite time from smooth initial conditions. A shockwave is a special case of a discontinuity that is forced by the dynamics: unlike the linear advection equation, the discontinuity can propogate without progressive diffusion.

\subsection{Kuramoto-Sivashinsky Equation}
The KS equation is written as
\begin{equation}
    \frac{\partial u}{\partial t} + \nu\frac{\partial^4 u}{\partial x^4} + \frac{\partial^2 u}{\partial x^2} + \frac{1}{2}\frac{\partial u^2}{\partial x} = 0,
\end{equation}
and is used as a toy problem for turbulent flame fronts. Its dynamics lead to chaotic spatiotemporal behavior \cite{hyman1986kuramoto}. Chaos for PDEs is analagous to dynamical systems, defined by a small perturbation in initial conditions drastically affecting the time-evolution of the system \cite{strogatz2001nonlinear}.

\section{Methodologies}
\subsection{Network Architecture}
\begin{figure*}[h!]\label{diagrams}
\centering
\includegraphics[width=1\textwidth]{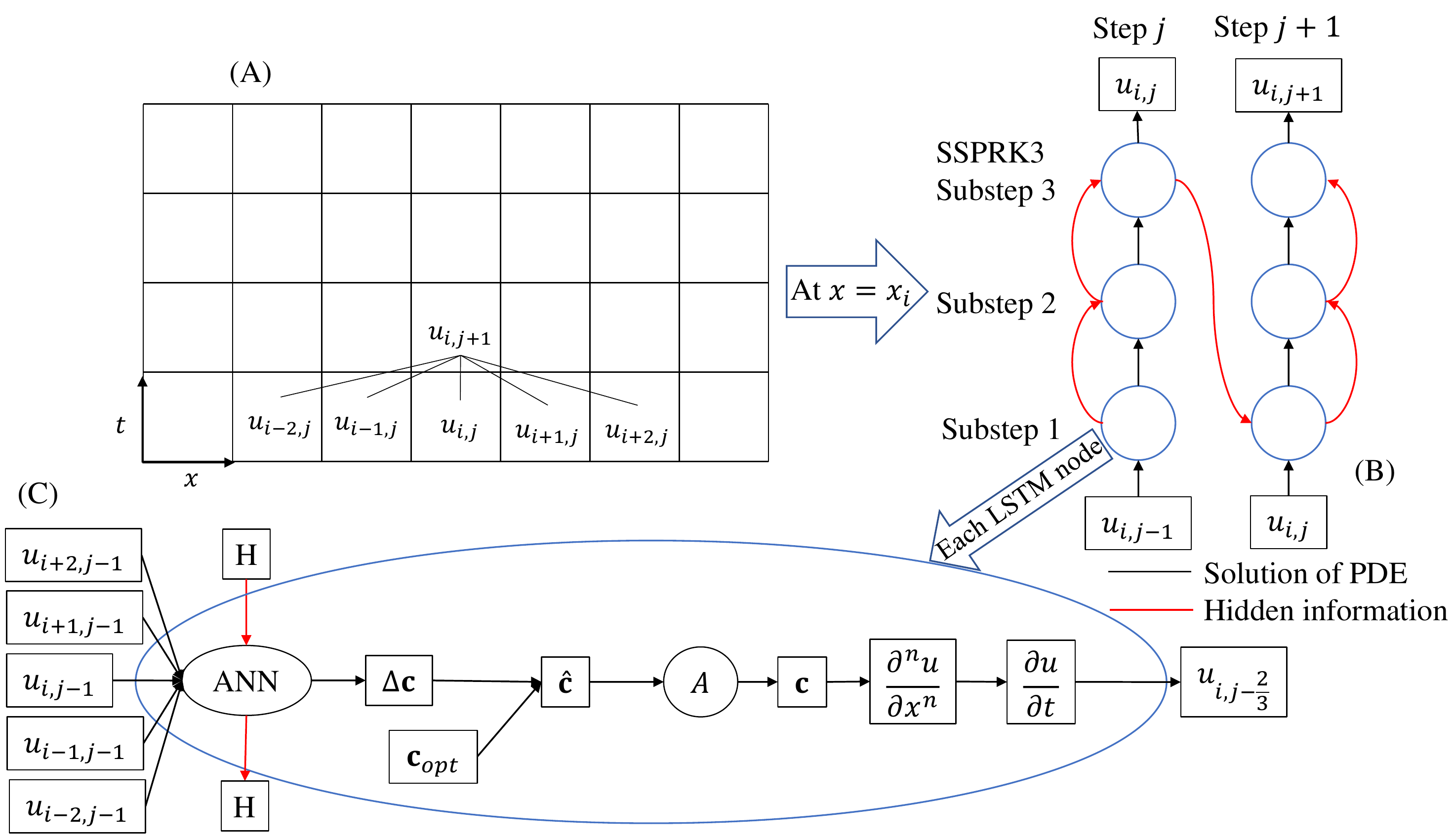}
\caption[batch1c]{Network Architecture at (A) the top level, (B)  LSTM at a specific $x$ location, (C) each evaluation of the LSTM}
\end{figure*}

We structure our network to utilize well-known methods from numerical PDEs described earlier (SSPRK, FDM, and FVM). The network architecture mimics the structure of a grid used to numerically solve a PDE, where the size of the filter of the convolutional layer corresponds to the stencil that selects which information is used to compute the derivative, and each output of the network corresponds to the solution of the PDE at time $t_j$ and location $x_i$. This can be seen in \ref{diagrams}.

When looking at a specific realization of the stencil in the $x$-domain, we arrive at an LSTM network. The purpose of the LSTM is to transfer information about the solution over long time-horizons, which adds a feature to our method that is not present in traditional methods. The hidden information is  transferred from substep to substep ($t_{j-1}$ to $t_{j-2/3}$ to $t_{j-1/3}$, and then to the next step at the end of the current step $t_{j-1/3}$ to $t_{j}$).

Within each evaluation of the LSTM, the information is used to compute the solution at the next substep in a way that mimics traditional FDM or FVM. The solution values at the previous timestep $u_{i-2:i+2,j-1}$ and the hidden information with dimension 32 from the previous timestep are input to a neural network with 3 layers and 32 neurons per layer. This network outputs the hidden information to the next substep, as well as a prediction of the FVM or FDM coefficients. The neural network does not directly predict these coefficients. Instead, it first computes a perturbation $\Delta \textbf{c}$ to the maximum-order coefficients $\textbf{c}_{opt}$, with $l_2$ regularization applied to $\Delta \textbf{c}$, similarly to a ResNet \cite{he2016deep}. We find that adding this step speeds up training and improves the performance of the model, as we are effectively biasing our estimate of the coefficients towards the coefficients that are optimal in the absence of prior knowledge about the solution. If the model is not confident about how perturbing $\textbf{c}_{opt}$ will affect the accuracy of the derivative it can output values close to zero to default back to $\textbf{c}_{opt}$. Once $\textbf{c}_{opt}$ has been perturbed by $\Delta \textbf{c}$ to obtain $\hat{\textbf{c}}$, an affine transformation is performed on $\hat{\textbf{c}}$ to guarantee that the coefficients are the desired order of accuracy \cite{bar2019learning}. This is one of the main benefits of using an FDM structure for our model, as we can prove that our model gives a solution that converges to the true solution at a known rate as the grid is refined. The details of this transformation can be seen in section 4.3.

After the affine transformation, the final coefficients $\textbf{c}$ are known. The spatial derivatives are computed by taking the inner product of the coefficients $c_{i-2:i+2}$ with $u_{i-2:i+2,j-1}$. For equations with multiple spatial derivatives, a different set of coefficients is computed for each. These spatial derivatives are then used to compute the time derivative, which can finally be used to compute the solution at the next substep. This process is then repeated for the desired number of timesteps. Our network architecture allows us to train on exact solutions or data of the PDE end-to-end.

\subsection{Training Algorithm}
We train our network in a way that exactly mimics how it would be used to solve a PDE. More specifically, we start with some random initial condition and use the network to step the solution forward in time, and compare the result to the exact solution. For the linear advection equation, the analytical solution is known for arbitrary initial conditions. For the inviscid Burgers' equation and KS equation, the same simulation is also carried out on a fine grid using a baseline numerical method, which results in a solution that can be considered approximately exact. The inviscid Burgers' equation is solved using WENO5 FVM \cite{jiang1996efficient}, and the KS equation is solved using fourth-order FDM. The loss is computed by downsampling the exact solution onto the grid of the neural network, averaging over the square error at every point in time and space. We found this training strategy to be far superior to training on only 1 time-step at a time, as our approach is capable of minimizing long-term error accumulation and training the network to be numerically stable. In terms of computational complexity, our method is identical to backpropogation through time (BPTT) \cite{werbos1990backpropagation}.

\begin{algorithm}[tb]
   \caption{Train FiniteNet}
   \label{alg:train}
\begin{algorithmic}
   \STATE Select number of epochs $n_e$
   \STATE Select minibatch size $n_m$
   \STATE Select time-horizon $T$
   \FOR {$i=1$ {\bfseries to} $n_e$}
   \STATE Set total MSE to 0
   \FOR {$j=1$ {\bfseries to} $n_m$}
   \STATE Select initial condition $u_j(x,0)$
   \STATE Determine exact solution $u^*_j$
   \STATE Initialize hidden information $H(0)$ to 0
   \FOR{$k=1$ {\bfseries to} $T$}
   \STATE Compute $u_j(x,t_k)$ and $H(t_k)$ with FiniteNet
   \ENDFOR
   \STATE Add MSE between $u_j$ and $u^*_j$ to total MSE
   \ENDFOR
   \STATE Compute gradient of simulation error w.r.t. FiniteNet parameters
   \STATE Update FiniteNet parameters with ADAM optimizer
   \ENDFOR
\end{algorithmic}
\end{algorithm}

\subsection{Accuracy Constraints}
FiniteNet is structured such that the numerical method is guaranteed to satisfy a minimum order of accuracy $n$, $e = o(\Delta x^n)$ \cite{bar2019learning}. One can perform a Taylor series expansion on the approximations of the form of \ref{FDM_eq} to obtain linear constraint equations that the coefficients must satisfy for the method to achieve a desired order of accuracy. We take the coefficients that the neural network outputs, $\hat{\textbf{c}}$, and find the minimal perturbation
$\Delta \textbf{c}$ that causes them to satisfy the constraint equations, which leads to the optimization problem

\begin{equation}
\begin{array}{rrclcl}
\displaystyle \min_{\Delta \textbf{c} \in R^5} & \multicolumn{2}{l}{\sum_{n=-2}^{2} (\Delta c_n)^2}\\
\textrm{s.t.} & {A(\hat{\textbf{c}}+\Delta \textbf{c})=\textbf{b}},\\
\end{array}
\end{equation}

which has analytical solution $\Delta \textbf{c}=A^T(AA^T)^{-1}(\textbf{b}-A\hat{\textbf{c}})$, which can be expressed as an affine transformation on $\hat{\textbf{c}}$, and can therefore be added to the network as a layer with no activation function and untrainable weights.

\section{Experimental Results}
\subsection{Summary}
We find that our method is capable of reducing the error relative to the baseline method for all three equations tested. These results show promise for generalization to other equations, as each PDE we examined has qualitatively different behavior. A table summarizing our results can be seen in \ref{sample-table}. The variation is due to averaging results from randomly generated initial conditions. The error ratio $e_r$ is computed as $e_{r} = \frac{e_{F}}{e_{B}}$ where $e_{F}$ is the FiniteNet MSE and $e_{B}$ is the MSE of the baseline method. We analyze $\log_{10}e_r$ because $\lim_{e_{F}\to 0}e_r\to0$ and $\lim_{e_{B}\to 0}e_r\to\infty$. Hence, one case of the baseline method outperforming FiniteNet could skew the average and standard deviation of $e_r$ significantly while the opposite would have very little effect, resulting in a bias. Additionally, the data empirically follows a log-normal distribution more closely than a normal distribution.
\begin{table}[t]
\caption{Mean and standard deviation of $\log_{10}e_r$ for each PDE}
\label{sample-table}
\vskip 0.15in
\begin{center}
\begin{small}
\begin{sc}
\begin{tabular}{lcccr}
\toprule
Equation & $\log_{10}e_r$ & Better? \\
\midrule
Linear Advection    & -0.27$\pm$ 0.08& $\surd$ \\
Inviscid Burgers' & -0.53$\pm$ 0.08& $\surd$ \\
Kuramoto-Sivashinsky   & -0.62$\pm$ 0.41& $\surd$ \\
\bottomrule
\end{tabular}
\end{sc}
\end{small}
\end{center}
\vskip -0.1in
\end{table}

Hyperparameters were tuned to optimize performance on the linear advection equation. These same hyperparameters were then used as starting points for the other equations and tuned as necessary. For example, the time-horizon used in training was changed from 100 to 200 for the KS equation, as this helped FiniteNet track chaotic solutions over longer time-horizons. Additionally, it was found that increasing the $l_2$ regularization constant from $\lambda=0.001$ to $0.1$ for the KS equations improved performance. The inviscid Burgers' equation was trained for 500 epochs instead of 400 because the error was still decreasing,

All data generation, training, and testing was carried out on a desktop computer with 32 GB of RAM and a single CPU with 3 GHz processing speed. Hence, by scaling the computing resources one could scale our method to more challenging problems. Our code and trained models can be found here: \url{https://github.com/FiniteNetICML2020Code/FiniteNet}, which includes links to our data.

\begin{figure}[h!]\label{Recovery}
\centering
\includegraphics[width=0.4\textwidth]{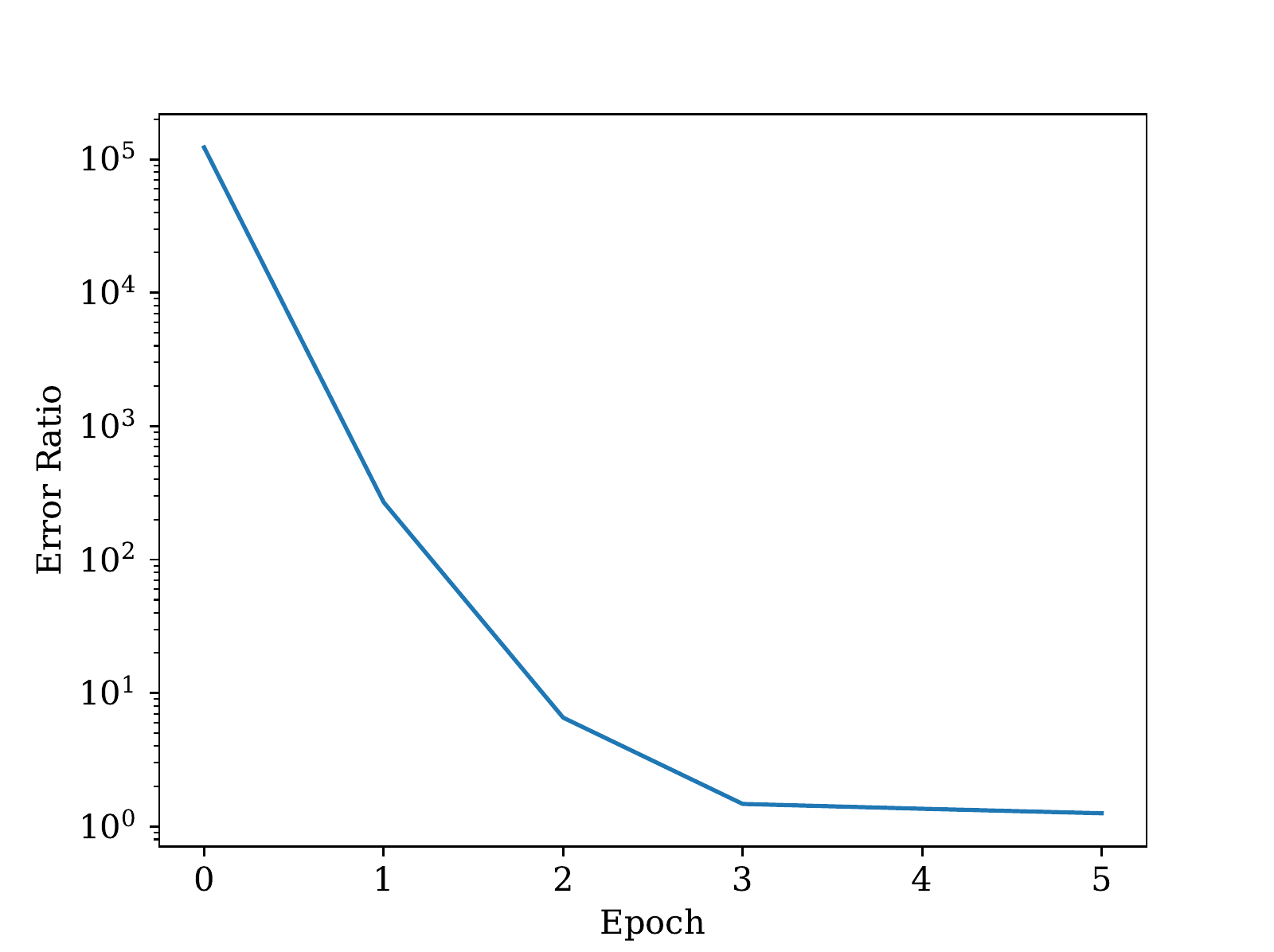}
\caption[batch1c]{Training FiniteNet from unstable initial condition}
\end{figure}

We also find that FiniteNet trains methods that are stable, which tends to be a challenge in the field of learned FDM. Although we cannot formally prove stability, we observe in \ref{Recovery} that if the randomly initialized network weights lead to an unstable scheme, the method will quickly learn to become stable by minimizing accumulated error.

\subsection{Linear Advection Equation}
For the linear advection equation, we generate random discontinuous initial conditions, and compare the errors obtained using FiniteNet to errors obtained using WENO5.  Each epoch involved generating 5 new initial conditions, computing the error against the exact solution, and updating the weights with the Adam optimizer \cite{kingma2014adam}. We trained for 400 epochs.

\begin{figure}[h!]\label{Adv_hist}
\centering
\includegraphics[width=0.4\textwidth]{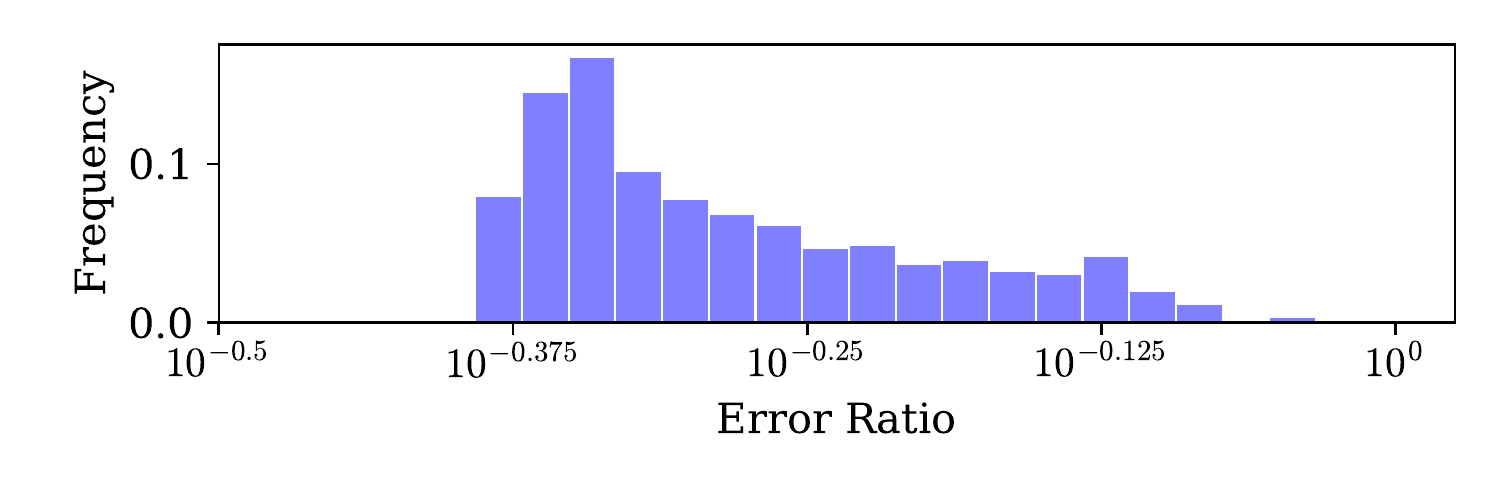}
\caption[batch1c]{PMF of linear advection testing}
\end{figure}

After training was complete, we tested the model on 1000 more random initial conditions and computed the error ratio. We saw that FiniteNet outperformed WENO5 in 999 of the 1000 simulations. A PMF of the error ratio can be seen in \ref{Adv_hist}.

\begin{figure}[h!]\label{adv_comp}
\centering
\includegraphics[width=0.4\textwidth]{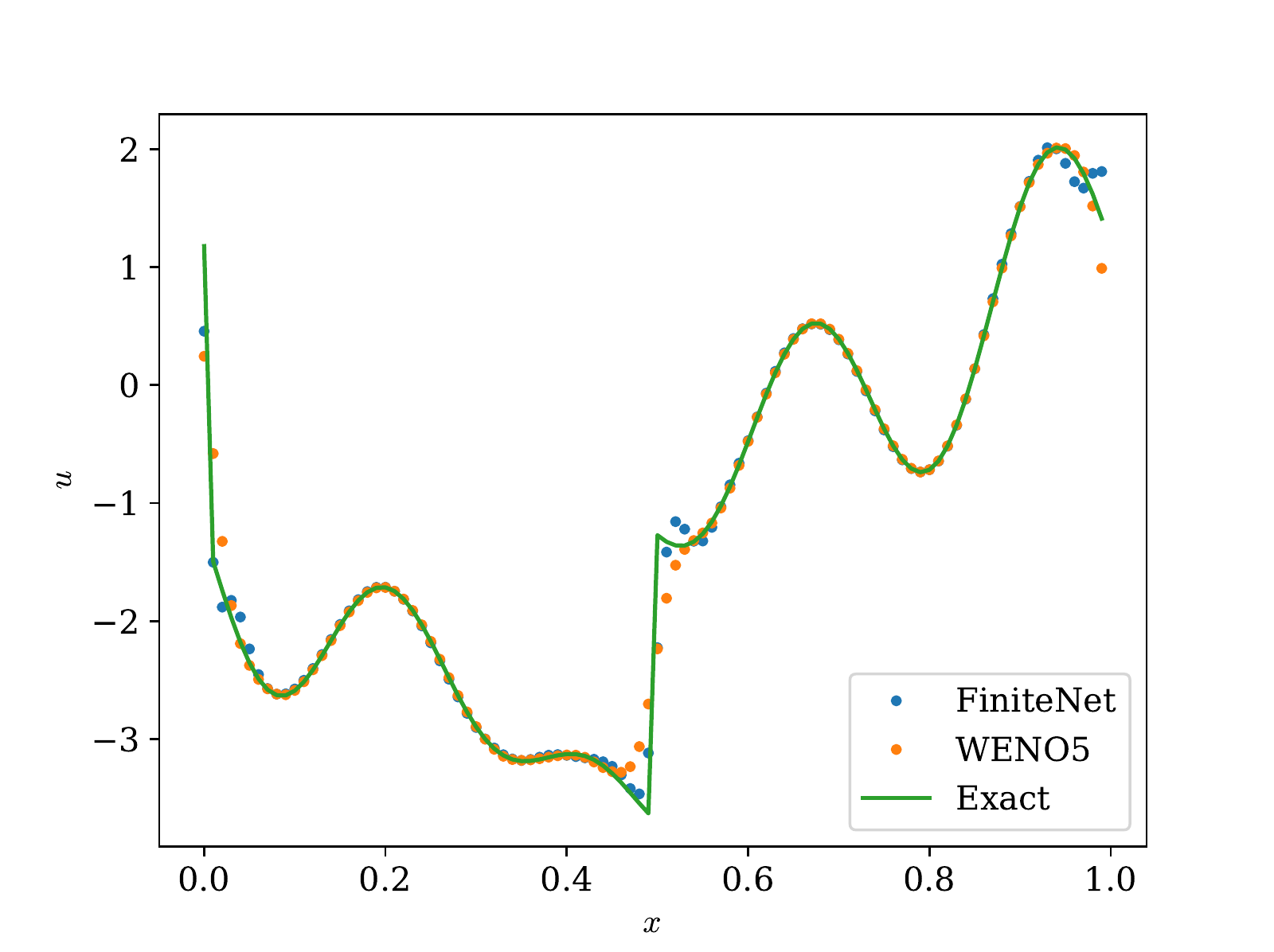}
\caption[batch1c]{Solutions of linear advection equation}
\end{figure}

We also plot a WENO5 solution and FiniteNet solution in \ref{adv_comp} to gain insight into how FiniteNet improves the solution. We see that FiniteNet more sharply resolves discontinuities at the cost of adding oscillations, which leads to a net reduction in error.

\begin{figure}[h!]\label{discVerr}
\centering
\includegraphics[width=0.4\textwidth]{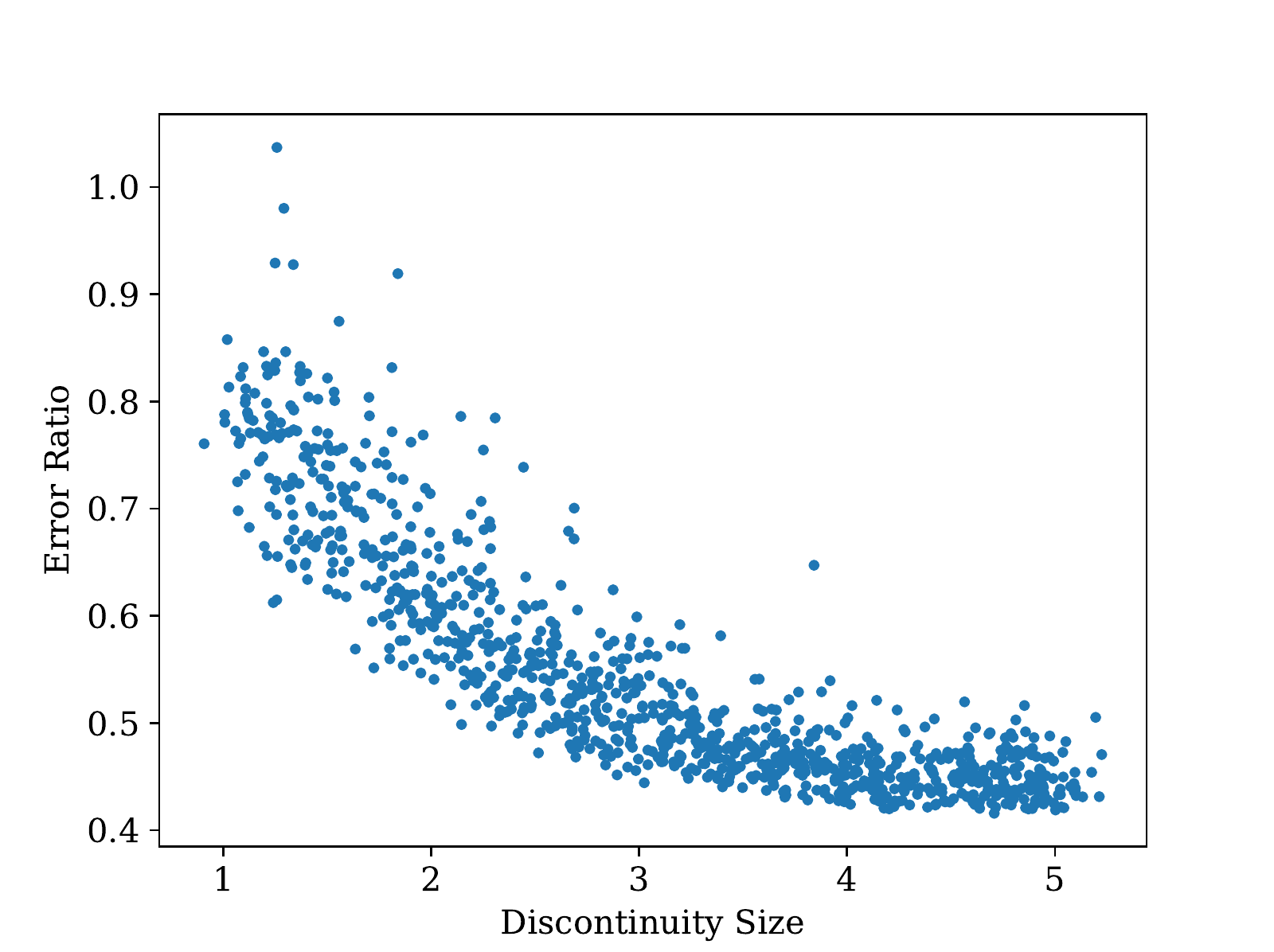}
\caption[batch1c]{Discontinuity size vs error ratio for advection equation}
\end{figure}

We can use this result to hypothesize that FiniteNet will further improve the solution when the discontinuity is larger. We verify our hypothesis by plotting error ratio against discontinuity width in \ref{discVerr}, and verifying that larger discontinuities lead to lower error ratios.

\subsection{Inviscid Burgers' Equation}
We train the neural network to interpolate flux values for the inviscid Burgers' equation. Once again, each epoch involves generating five random initial conditions. In lieu of an exact solution, we use WENO5 on a 4x refined mesh to approximate the exact solution. We then ran 1000 simulations with the trained network and compared the results to WENO5. FiniteNet achieved a lower error on 998 of the 1000 test cases.

\begin{figure}[h!]\label{burg_err_dat}
\centering
\includegraphics[width=0.4\textwidth]{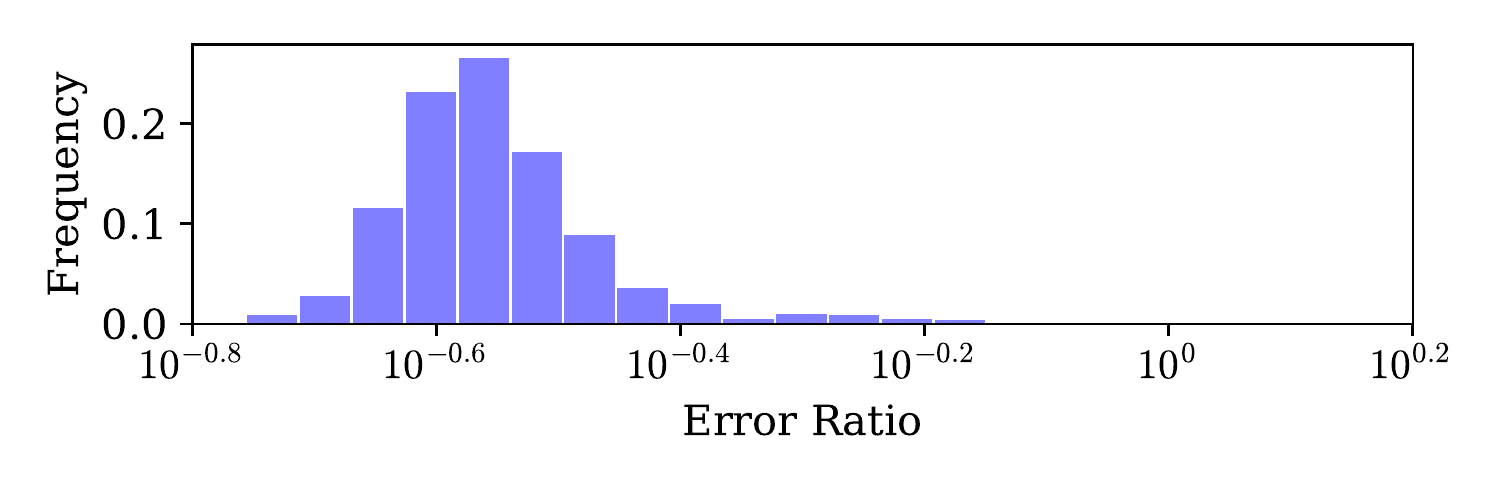}
\caption[batch1c]{PMF of inviscid Burgers' testing}
\end{figure}
By examining the error induced when numerically solving this equation with WENO5 and the FiniteNet, we see that WENO5 accumulates higher error around shocks. This tells us that FiniteNet achieves it's error reduction by more sharply resolving the shocks.

\begin{figure}[h!]\label{Burg_Err}
\centering
\includegraphics[width=0.4\textwidth]{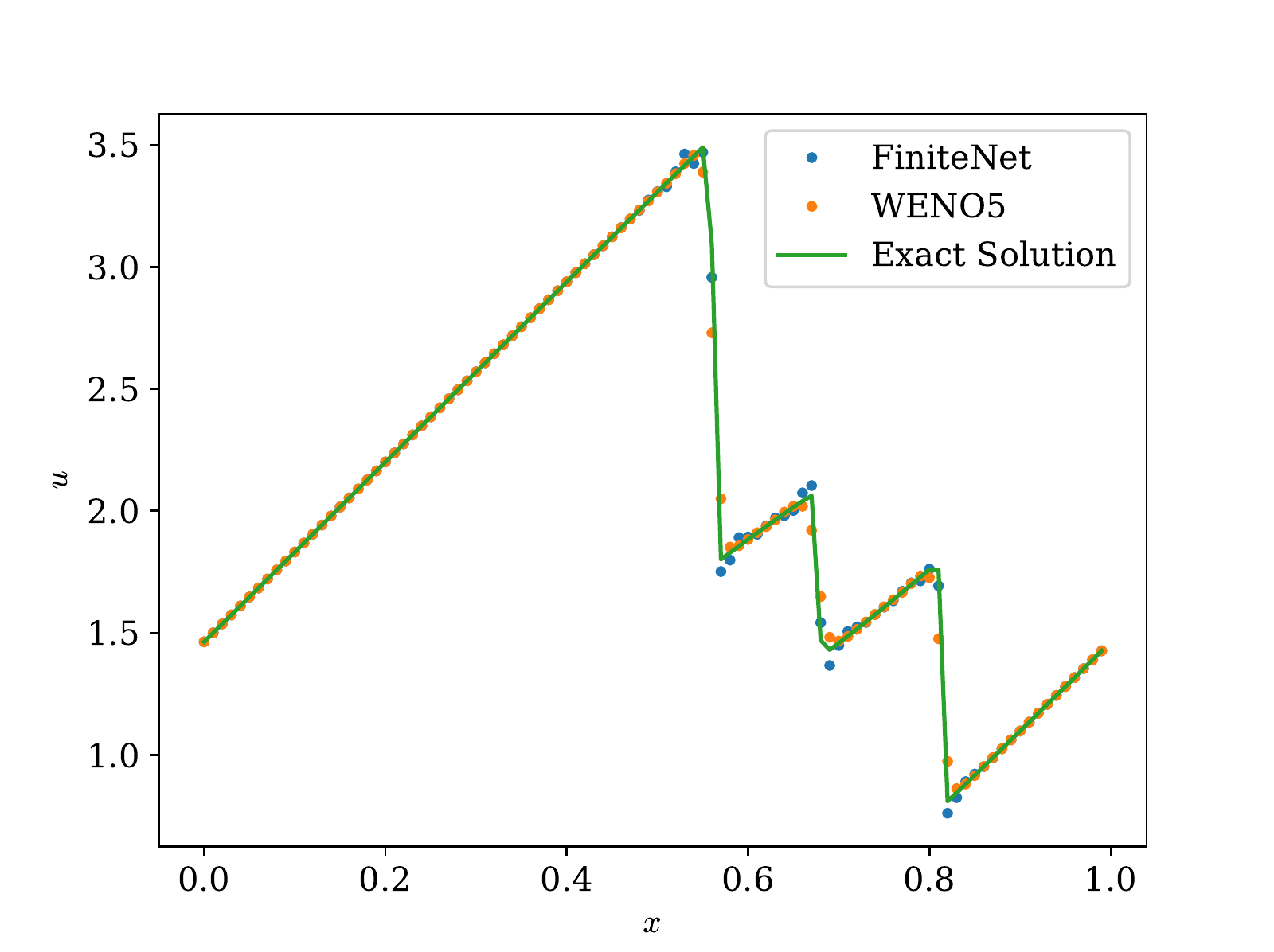}
\caption[batch1c]{Solution of inviscid Burgers' equation}
\end{figure}

\begin{figure}[h!]\label{TV_burg}
\centering
\includegraphics[width=0.4\textwidth]{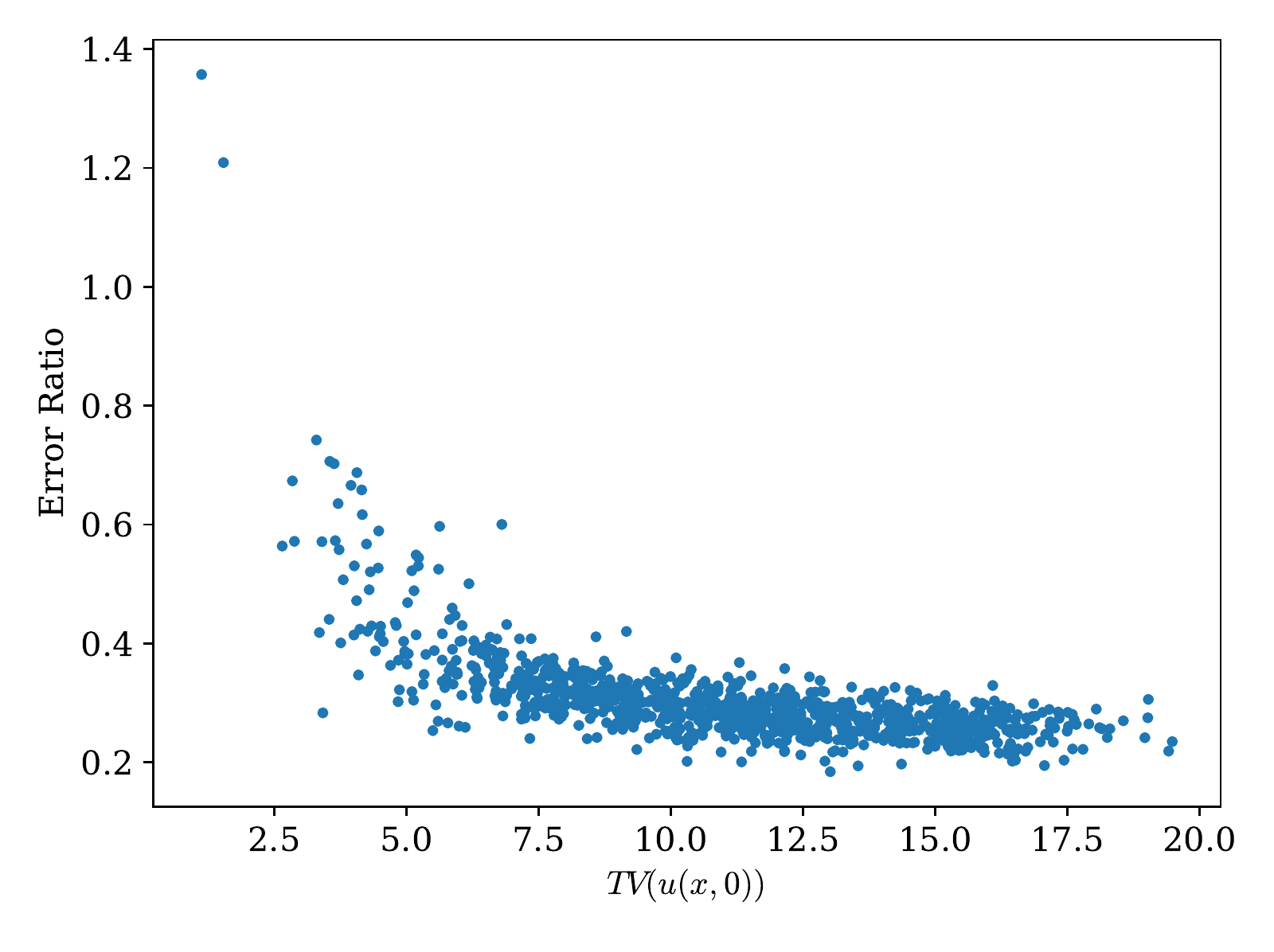}
\caption[batch1c]{Trends between total variation and error ratio}
\end{figure}

We can get a prediction of roughly how many and how large of shocks will develop in the solution by looking at the total variation of the initial condition. We plot the total variation of the initial condition and compare it to the error ratio in \ref{TV_burg}.

This data shows that FiniteNet tends to do better relative to WENO5 when the total variation of the initial condition is higher, which helps confirm our result that FiniteNet offers the most improvement when many large shocks form. 

\subsection{Kuramoto-Sivashinsky Equation}
\begin{figure*}[ht!]\label{KS_comp}
\centering
\includegraphics[width=1\textwidth]{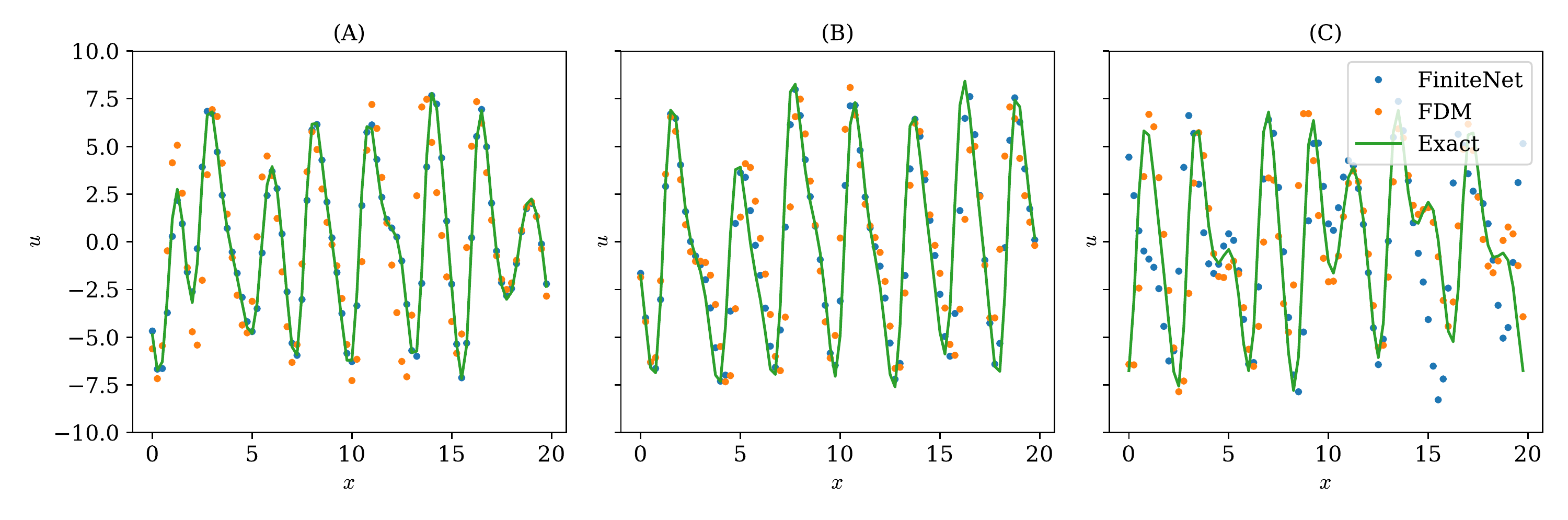}
\caption[batch1c]{(A) Best case, (B) typical case, and (C) worst-case error ratio between FiniteNet and FDM for KS}
\end{figure*}

The network is trained for the KS equation in the same way as was done for the inviscid Burgers' equation: by solving the equation on a 4x refined mesh to closely approximate the exact solution. We generate our random initial conditions for training by setting a random initial condition for the exact solution and simulating until the trajectory has reached the chaotic attractor, and start training from random sequences of the solution on the attractor so that we are learning a more consistent set of dynamics, as the initial transient tends to be less predictable.

After training, we use both FiniteNet and fourth-order FDM to solve the KS equations from 1000 new initial conditions. We find the FiniteNet achieves a lower error in 961 of the 1000 tests. Interestingly, we see that in some cases standard FDM is unable to track the chaotic evolution and results in a solution with no visual fit, while FiniteNet succeeds at tracking the chaotic trajectory. In the cases where FiniteNet leads to a larger error than FDM, we see that neither method could track the chaotic trajectory, so there is some small probability that FiniteNet may follow a worse trajectory than FDM. In order to determine the reliability of FiniteNet compared to FDM, we compute the statistics of the errors individually. FiniteNet has mean error 1.20 and standard deviation 1.89, while FDM has mean error 2.94 and standard deviation 3.01. So we see that FDM error has higher mean and variance, and can conclude that FiniteNet is more reliable.

\begin{figure}[ht!]\label{ks_hist}
\centering
\includegraphics[width=0.4\textwidth]{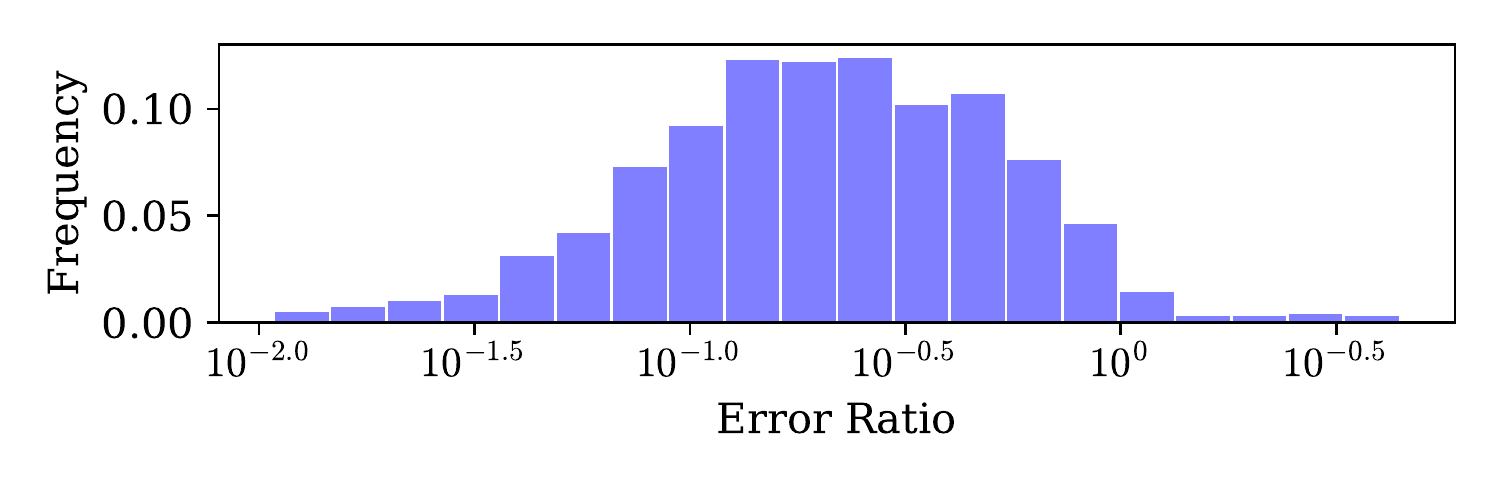}
\caption[batch1c]{PMF of KS testing}
\end{figure}
\section{Conclusions}
%Solving PDEs remains a challenge in the fields of physics and engineering. 
In this paper, we have presented FiniteNet, a machine learning approach that can reduce the error of numerically solving a PDE. %\tim{You didn't demonstrate this, you showed that it was more accurate!} \ben{that's true, I updated the sentence accordingly}
By combining the LSTM with well-understood and tested discretization schemes, we can significantly reduce the error for PDEs displaying a variety of behavior including chaos and discontinuous solutions. The FiniteNet architecture mimics the structure of a numerical PDE solver, and builds the timestepping method, spatial discretization, and PDE into the network. We train FiniteNet by using it to simulate the PDE, and minimizing simulation error against a trusted solution. This training approach causes the resulting numerical scheme to be empirically stable, which has been a challenge for other approaches.

We compared numerical solutions obtained by FiniteNet to those of baseline methods. We saw that for the inviscid Burgers' equation and the linear advection equation, FiniteNet is more sharply resolving discontinuities at the cost of sometimes adding small oscillations. This result makes intuitive sense, as the global error is dominated by regions near discontinuities so FiniteNet reduces the error the most by improving performance in these areas. When examining the KS equations, we see that FiniteNet reduces the error by more accurately tracking the chaotic trajectory of the exact solution. The main challenge of chaotic systems is that small errors grow quickly over time, which can lead to solutions that have completely diverged from the exact solution. FiniteNet is able to significantly reduce error by preventing this from happening in many realizations of this PDE.

Further work could involve comparing the runtime vs. error of FiniteNet to baseline approaches to more directly characterize how much of an improvement FiniteNet can offer. Additionally, we have till now only examined problems in one spatial dimension with periodic boundary conditions. %and do not address other types commonly seen in practical problems such as inflows or walls. Lastly, 
It will be interesting to test the method on a large-scale problem, such as a turbulent flow.
\section*{Acknowledgements}
This material is based upon work supported by the National Science Foundation Graduate Research Fellowship under Grant No. 1745301
\bibliography{example_paper}
\bibliographystyle{icml2020}

\end{document}